\documentclass[manuscript,screen]{acmart}

\AtBeginDocument{%
  }
\usepackage[caption=false,font=footnotesize,labelfont=rm,textfont=rm]{subfig}
\usepackage{bm}
\usepackage{graphicx}
\usepackage{float}
\usepackage{makecell}
\usepackage{tabularx}
\usepackage[utf8]{inputenc}

\DeclareUnicodeCharacter{2113}{\ensuremath{\ell}}

\acmISBN{978-1-4503-XXXX-X/18/06}

\begin{document}

\title{Knockoff-Guided Feature Selection via A Single Pre-trained Reinforced Agent}


\author{Xinyuan Wang}
\affiliation{
  \institution{School of Computing and Augmented Intelligence, Arizona State University, Tempe}
  \country{USA}}
\email{xwang735@asu.edu}

\author{Dongjie Wang}
\affiliation{%
  \institution{Department of Computer Science, University of Kansas, Lawrence}
  \country{USA}
}\email{wangdongjie@ku.edu}

\author{Wangyang Ying}
\affiliation{
  \institution{School of Computing and Augmented Intelligence, Arizona State University, Tempe}
  \country{USA}}
\email{yingwangyang@gmail.com}

\author{Rui Xie}
\affiliation{%
  \institution{Department of Statistics and Data Science, University of Central Florida, Orlando}
  \country{USA}}
\email{Rui.Xie@ucf.edu}

\author{Haifeng Chen}
\affiliation{%
  \institution{NEC Laboratories America Inc, Princeton}
  \country{USA}}
\email{haifeng@nec-labs.com}

\author{Yanjie Fu}
\affiliation{%
  \institution{School of Computing and Augmented Intelligence, Arizona State University, Tempe}
  \country{USA}}
\email{yanjie.fu@asu.edu}

\renewcommand{\shortauthors}{Wang et al.}

\begin{abstract}
    Feature selection prepares the AI-readiness of data by eliminating redundant features.
    Prior research in this area falls into two primary categories:
    i) Supervised Feature Selection (SFS), which identifies the optimal feature subset based on their relevance to the target variable; ii) Unsupervised Feature Selection (UFS), which reduces the feature space dimensionality by capturing the essential information within the feature set instead of using target variable.
    However, SFS approaches suffer from time-consuming processes and limited generalizability across different scenarios due to the dependence on the target variable and downstream ML tasks. 
    UFS methods are constrained by the 
    deducted feature space is latent and untraceable.
    To address these challenges, we introduce an innovative framework for feature selection, which is guided by knockoff features and optimized through reinforcement learning, to identify the optimal and effective feature subset.
    In detail, our method involves generating "knockoff" features that replicate the distribution and characteristics of the original features but are independent of the target variable. Each feature is then assigned a pseudo label based on its correlation with all the knockoff features, serving as a novel metric for feature evaluation.
    Our approach utilizes these pseudo labels to guide the feature selection process in three novel ways, optimized by a single reinforced agent:
    1). A deep Q-network, pre-trained with the original features and their corresponding pseudo labels, is employed to improve the efficacy of the exploration process in feature selection.
    2). We introduce unsupervised rewards to evaluate the feature subset quality based on the pseudo labels and the feature space reconstruction loss to reduce dependencies on the target variable.
    3). A new $\epsilon$-greedy strategy is used, incorporating insights from the pseudo labels to make the feature selection process more effective.
    Finally, extensive experiments have been conducted to illustrate the superiority of our framework. The code and data are publicly available at the \href{https://github.com/anord-wang/KGPTFS.git}{provided link}.
\end{abstract}

\ccsdesc[500]{Computing methodologies~Artificial intelligence}



\keywords{Feature Selection, Reinforcement Learning, Unsupervised Reward, Knockoff, Matrix Reconstruction}


\maketitle

\section{Introduction}
Feature selection aims to identify the optimal feature subset by eliminating redundant and irrelevant features.
This process can lead to enhanced predictive accuracy, lower computational costs, improved explainability, and faster deployment of machine learning models~\cite{li2017feature,miao2016survey}.
It has been effectively adopted in various environments, including biomarker identification, intrusion detection, urban computing, financial analysis, etc.

Traditional feature selection methods commonly can be divided into two main categories: Supervised Feature Selection (SFS), which relies on the target variable to evaluate and select features, and Unsupervised Feature Selection (UFS), which aims to capture the essential information within the feature set, independent of the target variable.
Despite the wide application of feature selection in theory and practice, existing methods, including filter~\cite{da2009weighted} ~\cite{wang2017feature}, wrapper~\cite{granitto2006recursive}~\cite{whitney1971direct}, and embedded ~\cite{tibshirani1996regression}~\cite{sugumaran2007feature} methods, still face significant challenges. Each method has its advantages, but also notable drawbacks. Firstly, model-dependent feature selection methods may not consider the global perspective, potentially leading to local optimum solutions and an inability to automatically update with changing environmental states. Secondly, the reliance on dataset labels by current feature selection methods significantly narrows their applicability, increasing computational costs and time.

\begin{figure}[htbp]
\centering
\includegraphics[width=0.5\textwidth]{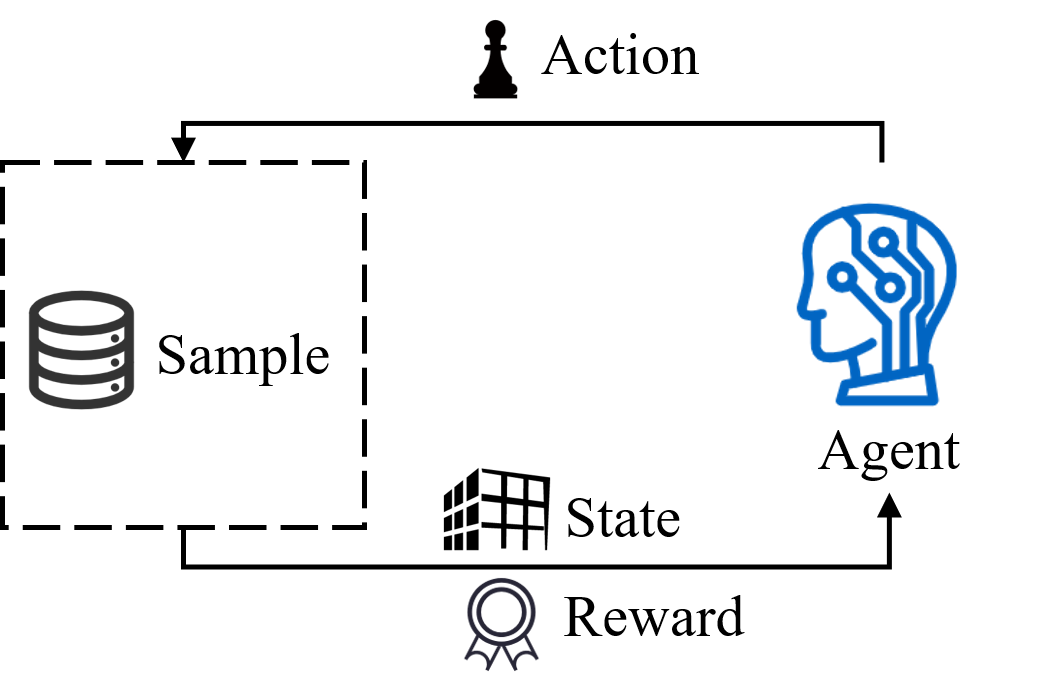}
\caption{The Unsupervised Feature Selection Structure.}
\label{fig:unsupervised}
\end{figure}


To overcome these challenges, we propose a novel feature selection framework based on the generation of "knockoff" features and reinforcement learning techniques. Recently, reinforcement learning (RL) methods have been introduced into the domain of feature selection. Model-free RL can achieve optimal long-term decisions in unknown environments. These advantages allow RL to automate the exploration of feature subspaces globally, achieving better performance than traditional methods. In this paper, we develop an automated unsupervised feature selection method. For the automated part, our framework employs reinforcement learning, specifically the DQN algorithm. For the unsupervised part, we utilize the Knockoff method to generate pseudo information and Auto-Encoders to construct matrix reconstruction rewards, replacing label-based feedback. This structure, as shown in Figure~\ref{fig:unsupervised}, does not require downstream labels of the dataset. In this unsupervised approach, the environment focuses solely on the data itself.

The automated approach can expedite the process of feature selection, which is highly advantageous in today's era of data explosion. Using reinforcement learning to allow the agent to choose features autonomously is a promising method. Unsupervised methods have two advantages: 1) they can accelerate the feature selection process as there is no need for time-consuming downstream tasks for validation after each decision. 2) they can broaden the applicability of the method, as datasets with perfect labels are scarce, and the majority are unprocessed datasets. Unsupervised methods are equally applicable to these raw datasets. However, unsupervised methods also have two drawbacks: 1) how to guide the agent in feature selection in the absence of labeled data? 2) how to provide feedback on the agent's decisions without engaging in downstream tasks? These two issues are the challenges addressed in this paper.

To be more specific, the first challenge is to get unsupervised knowledge from the data to guide feature selection. Incorporating data information into the feature selection step has always been a challenge. It can help the model make better decisions based on the current data. The current black box structure (neural network) is not very interpretable, and the network structure is overly dependent on the domain and structure of datasets~\cite{buhrmester2021analysis}.
The second challenge is to evaluate selection results without data labels. The goal of feature selection is to select a feature subset suitable for downstream tasks, so performance on downstream tasks is a perfect way to evaluate the current feature subset. Unfortunately, it requires completing a downstream task after each selection, which will greatly lengthen the training time, especially when the number of features is very large. At the same time, this is still highly dependent on data labels. However, in the current big data era, the number of labeled datasets is small and the cost is high. Therefore, how to design a fast, unsupervised evaluation method is a challenge. At the same time, it is also necessary to ensure that the designed evaluation method can truly reflect the information richness of the feature subset to ensure performance in downstream tasks.

To achieve the goal of automated unsupervised feature selection, we use a Deep Q-Learning Network(DQN) as the overall framework of the solution~\cite{liu2019automating}. A single agent is used to select each feature to obtain the final feature subset~\cite{zhao2020simplifying}. After multiple rounds of iterations, we obtain the final feature subset, which is used as the final selected feature. Model-free reinforcement learning can obtain long-term optimal decisions in unknown environments by interacting with the dynamic environment, maintaining the exploration-exploitation trade-off, and receiving reward signals as feedback~\cite{sutton2018reinforcement}. These advantages give reinforcement learning the ability to globally automate feature subspace exploration, thereby achieving better performance than traditional methods. 
For the first challenge, we employ the Knockoff method to generate the pseudo information to guide the decision process in the reinforcement learning process, which can generate a feature set that is highly similar to the input data but not related to the data label~\cite{candes2018panning}. This unique information matrix, through its mathematical characteristics, provides additional information in an unsupervised manner, indirectly aiding the agent in making selections. Feature labels are obtained by comparing the relationship between features and ‘Knockoff’ features. Through these labels, the decision-making network can be pre-trained in advance and $\varepsilon$-greedy strategy can be guided. 
For the second challenge, we evaluate the agent's decision with a novel unsupervised reward consisting of feature labels and matrix reconstruction without the support of downstream tasks. First, we employ two Auto-Encoders to measure the information richness difference between the original feature set and the selected subset. Secondly, we compare the selection results of the decision-making network and the Knockoff labels to measure the accuracy of DQN's judgment of the vector. Finally, feature redundancy evaluation is added to prevent redundant features from being retained. 
To verify the effectiveness of our proposed structure, we tested it on multiple datasets of different task types. The datasets we used include classification tasks and regression tasks, cover data in different fields, and also have a large number of features and samples. 

Our contributions are shown below:
\begin{itemize}
  \item \textbf{Automated Unsupervised Feature Selection Framework}: We propose a novel framework utilizing reinforcement learning, specifically designed for automated unsupervised feature selection. Leveraging the Knockoff method for pseudo information and Auto-encoder for matrix reconstruction, our approach transcends the constraints of supervised information and reward structures.
  \item \textbf{Integration of Knockoff Information}: We introduce and integrate Knockoff information into the reinforcement learning process, enhancing interpretability and providing a mechanism to guide decision-making during the feature selection process.
  \item \textbf{Novel Reward Function Construction}: To overcome challenges related to the absence of labeled data and downstream tasks, we construct a new reward function using matrix comparison and Knockoff labels. 
  \item \textbf{Empirical Validation}: We conduct extensive experiments on diverse datasets, including classification and regression tasks in various fields. Through contrast and ablation experiments, we demonstrate the feasibility and effectiveness of our proposed structure, showcasing its automation, unsupervised nature, and generalization capabilities across different datasets and task types.
\end{itemize}

\section{Methodology} \label{Methodology}

\subsection{Problem Definition}

In this article, we implement feature selection tasks in an automated, unsupervised manner. Therefore, the machine learning task of this article is feature selection, which is to automatically select a subset from the original feature set without relying on data labels to better meet the needs of downstream tasks. A given dataset ${\bm{D}}$ including n samples, where each sample includes d-dimensional features, so the dataset can be shown as ${\bm{D}}=\left(f_1,f_2,\ldots,f_d\right)^T$. For this dataset, we build a feature selection method based on reinforcement learning to obtain the selected subset ${\bm{D}}_{sub}=\left(f_i,f_j,\ldots,f_k\right)^T$, where $ \bm{D}_{sub} \in \bm{D} $, which is used to meet downstream tasks need while reducing the amount of calculation.

\subsection{Framework Overview}
\begin{figure}[htbp]
\centering
\includegraphics[width=1.0\linewidth]{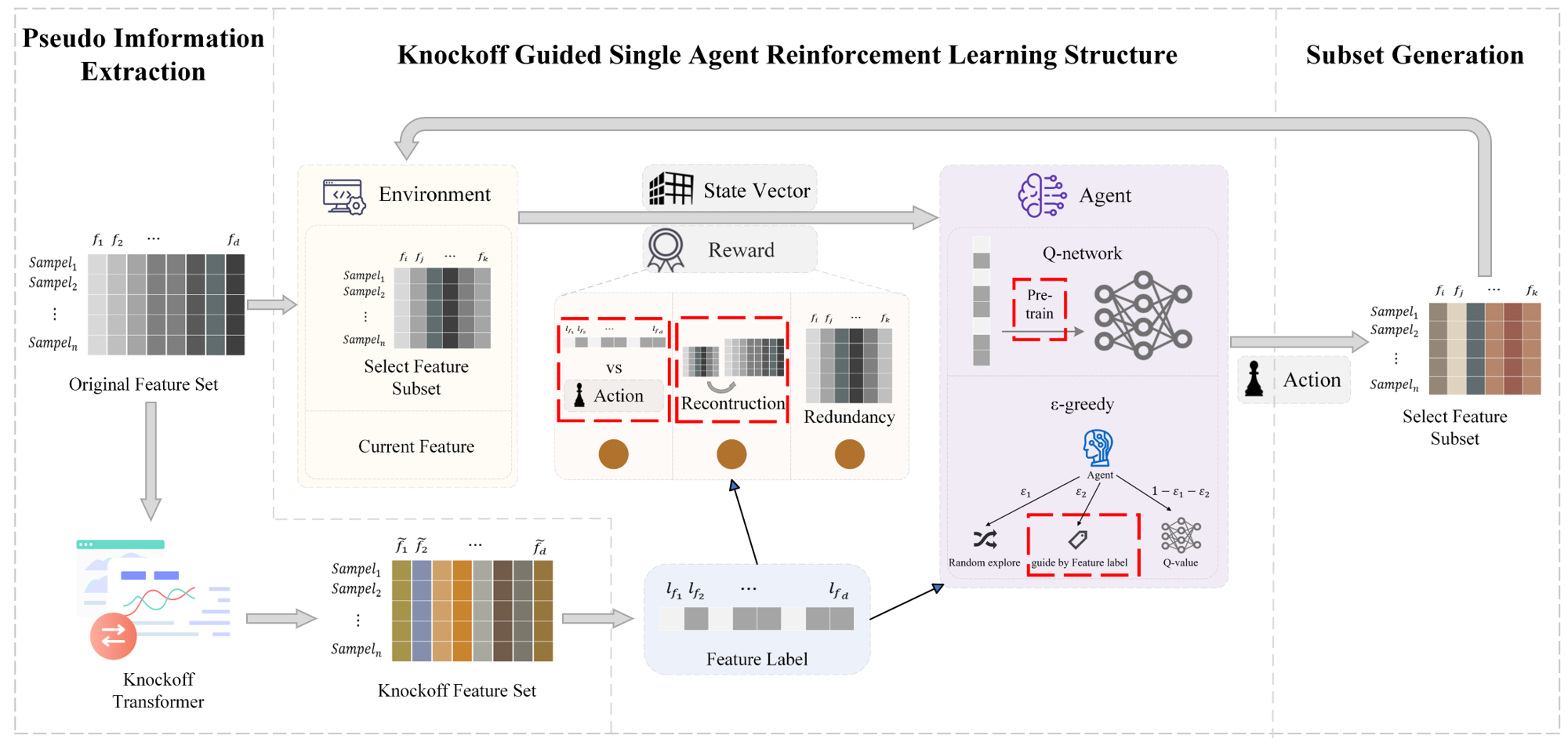}
\caption{The Overall Structure. We employ a single agent to make decisions for each feature. Utilizing the Knockoff information, along with an autoencoder, our innovative structure incorporates four key elements: 1) Pre-training: We pre-train the decision network of the Deep Q-Network using feature labels to guide the agent in making decisions for each feature. 2) $\varepsilon$-Greedy Policy: In addition to random exploration, we use feature labels to guide the agent in the $\varepsilon$-greedy policy. 3) Knockoff Reward: We compare the agent's decisions with feature labels to generate an unsupervised reward function. 4) Matrix Reconstruction Reward: We obtain representation vectors for both the original feature set and the feature subset, calculating an unsupervised matrix reconstruction reward.}
\label{fig:overall}
\end{figure}

To solve the above feature selection problem, we propose an automated unsupervised feature selection structure based on reinforcement learning. More specifically, we employ a single agent DQN algorithm, as shown in Figure~\ref{fig:overall}.

In this framework, the agent will make decisions on each feature in sequence with the help of injected pseudo information, which is obtained by the Knockoff matrix, and the obtained new feature subset is used as the next state. Then we build an unsupervised reward function and perform reinforcement learning training.
We employ the Knockoff matrix as pseudo information for the feature set to be processed. The Knockoff method can generate a pseudo feature set that is the same distribution as the sample but not related to the label, which is unsupervised. By comparing the relationship between each feature and this pseudo feature set, we can generate feature labels based on the pseudo information. This label is used in the reinforcement learning decision network pre-training stage and the initialization of the $\varepsilon$-greedy policy to complete pseudo information injection.
There are two parts of unsupervised information to form the reward function. First, the matrix reconstruction is used to obtain the difference between the selected feature subset and the entire feature set as part of the reward function. Then another part of the reward function is constructed based on comparing the agent action and the feature labels obtained through the Knockoff matrix. This new unsupervised reward function is used to replace the original supervised reward function that is mainly based on downstream task performance.
The following sections will be a detailed introduction to the three parts.

\subsection{Single Agent Reinforcement Learning Based Feature Selection Framework}

Using the DQN algorithm, the agent makes judgments on features in order. First, for the feature set ${\bm{D}}=\left(f_1,f_2,\ldots,f_d\right)^T$, the Model-X Knockoff algorithm~\cite{candes2018panning} is employed to construct a pseudo feature set $\widetilde{\bm{D}}=\left({\widetilde{f}}_1,{\widetilde{f}}_2,\ldots,{\widetilde{f}}_d\right)^T$, Feature labels based on pseudo information are then obtained and used for pre-training of the DNQ decision network and the initialization of the $\varepsilon$-greedy policy, which will be introduced in detail in section~\ref{Pseudo Information Injection}. Each important component of reinforcement learning will be introduced next.

\textbf{Single agent}: There is only one agent in this framework, which greatly reduces the number of model parameters. Given the current subset of features, this agent is responsible for making decisions for the next feature in sequence. It also receives rewards as feedback.

\textbf{Environment}: The environment is a key factor in reinforcement learning, interacting with the agent and providing rewards as feedback. In this article, the selected feature subset and the current feature index to be processed constitute the environment. Every time the agent takes an action, the environment will change, whether it is the feature subset or the current feature index.

\textbf{State}: The agent can determine the optimal action by observing the current environmental state. In this article, state s is used to describe the selected feature subset and the current feature index. To extract the representation of s, we use a convolutional Auto-Encoder for state description. We first use the feature subset for self-supervised training. By utilizing the encoder and decoder of the Auto-Encoder to reconstruct the feature subset, the Auto-Encoder can capture the underlying information embedded in it. This information is reflected in the representation vectors obtained through the encoder. In this way, we can obtain a well-crafted representation vector for the feature subset. The training process of the Auto-Encoder is as follows:
\begin{equation}
  {\bm{D}}_{rec}=Decoder\left(Encoder\left({\bm{D}}_{sub}\right)\right)
\end{equation}
We employ the L2 norm as the loss function for the self-supervised training of the Auto-Encoder: 
\begin{equation}
\text{loss}_{AE} = \|\bm{D}_{rec} - \bm{D}_{sub}\|_2^2
\end{equation}

Then we use the encoded vector $Encoder\left({\bm{D}}_{sub}\right)$ as the state representation vector. For the current feature index, we incorporate it into the state space in the form of a one-hot encoding vector.

\textbf{Action}: The action in this framework is to make a judgment on the current feature. The action space includes two situations $A=\left\{0,1\right\}$, where $a=1$ means selecting the current feature, and $a=0$ means not selecting the current feature.

\textbf{Reward}: Rewards give the agent feedback about its actions based on the current state of the environment. In this article, we use Pseudo Information and Matrix Reconstruction to jointly construct the reward function. The final reward function is the sum of three parts, shown as:
\begin{equation} \label{reward}
  R=R_{mr}\left({\bm{D}}_{sub}\right)+R_{pi}\left(f_i\right)+R_{rd}\left(f_i\right)
\end{equation}

For the currently selected feature subset, we use Matrix Reconstruction to compare the difference with the entire feature set and construct the reward function, which is shown as $R_{mr}\left({\bm{D}}_{sub}\right)$ in Equation ~\ref{reward}. 
For the current candidate feature, compare the action given by the agent and the feature labels based on the pseudo information, and construct the reward function, which is shown as $R_{pi}\left(f_i\right)$ in Equation ~\ref{reward}. 
At the same time, to avoid selecting too many redundant features, we add the global redundancy measure of features into the final reward function, which is shown as $R_{rd}\left(f_i\right)$ in Equation ~\ref{reward}.
The new unsupervised reward function will be detailed in section~\ref{Unsupervised Reward}.

\subsection{Unsupervised Pseudo Information Injection to Guide the Agent's Decision} \label{Pseudo Information Injection}

In reinforcement learning, the agent makes decisions by observing the environment, and actions will change the environment and rewards will be generated to provide feedback to the agent. Therefore, the perception of the environment is extremely important, and in the proposed framework, the dataset is an important part of the environment. We are going to help the agent make better decisions and speed up the reinforcement learning training process by generating augmented information about the dataset. 

Most of the existing methods focus on the data themselves, using neural networks to encode the data and obtain information representation vectors. However, this black-box approach cannot penetrate information into different parts of the reinforcement learning structure and has limited interpretability. Therefore, we introduced the tool Model-X Knockoff to extract data information. The Knockoff method uses noise variables (knockoff methods) to enhance the observed data. It can generate a feature set that is highly similar to the input data but not related to the data label~\cite{candes2018panning}. By comparing the relationship between the features to be judged and Knockoff features, we can get labels for each feature. Through these labels, we can pre-train the decision-making network in the DQN structure and complete the initialization with a guiding role. Compared with random initialization, our method can provide preliminary evaluation criteria for the decision-making network and accelerate the convergence of the network. At the same time, in the $\varepsilon$-greedy policy in the beginning stage, these labels are used to replace a part of the random selection, so that the entire reinforcement learning training process converges faster.

Model-X Knockoff algorithm is a feature selection method proposed by E. J. Candès et al., which is used for variable selection and control of false discovery rate (FDR) in the context of high-dimensional data analysis~\cite{dai2022false}. The Model-X Knockoff method is designed to solve the problem of multiple hypothesis testing when dealing with a large number of variables. The main idea is to create a set of ‘fake’ variables that mimic the original variables while ensuring that they are independent of the response variable. These imitation variables can serve as a baseline against which to compare the importance of the original variables. That is, the Knockoff matrix $\widetilde{{D}}=\left({\widetilde{f}}_1,{\widetilde{f}}_2,\ldots,{\widetilde{f}}_d\right)^T$ is generated for the data ${D}=\left(f_1,f_2,\ldots,f_d\right)^T$, which has two characteristics:

\begin{itemize}
\item Identical distribution: For any subset S, $\left({D},\widetilde{{D}}\right)_{swap\left(S\right)}=\left({D},\widetilde{{D}}\right)$. In this context, $swap$ denotes the interchange of any feature between ${D}$ and $\widetilde{{D}}$. This characteristic signifies that Knockoff features share the same distribution as the original features, ensuring that the collection remains equivalent even after exchanging features.
\item Independent of $Y$, and it is also guaranteed even though Knockoff matrix $\widetilde{D}$ is constructed without looking at $Y$.
\end{itemize}

Here we use Knockoff because it can enhance the data from negative aspects and it is unsupervised. These two properties fit our model very well. Based on the relationship between each feature and this pseudo feature set $\widetilde{D}$, we construct a label for each feature. These feature labels can reflect the quality of the feature to a certain extent. Obviously, features that are close to the pseudo feature set that have nothing to do with $Y$ are not the mainstay in the downstream task of predicting $Y$. However, we cannot assume that features far away from the pseudo feature set will make outstanding contributions to downstream tasks. Therefore, we use this part of pseudo information, but do not regard it as a strong rule in reinforcement learning. These labels are used to pre-train the QDN decision network and replace random selection in the $\varepsilon$-greedy strategy, while also contributing to the unsupervised reward function.

\subsubsection{DQN Decision-Making Network Pre-train} \label{Pre-train}
First, we use these feature labels and features to pre-train the DQN decision network, which is performed through supervised learning. 

\begin{figure}[htbp]
\centering
\includegraphics[width=0.5\linewidth]{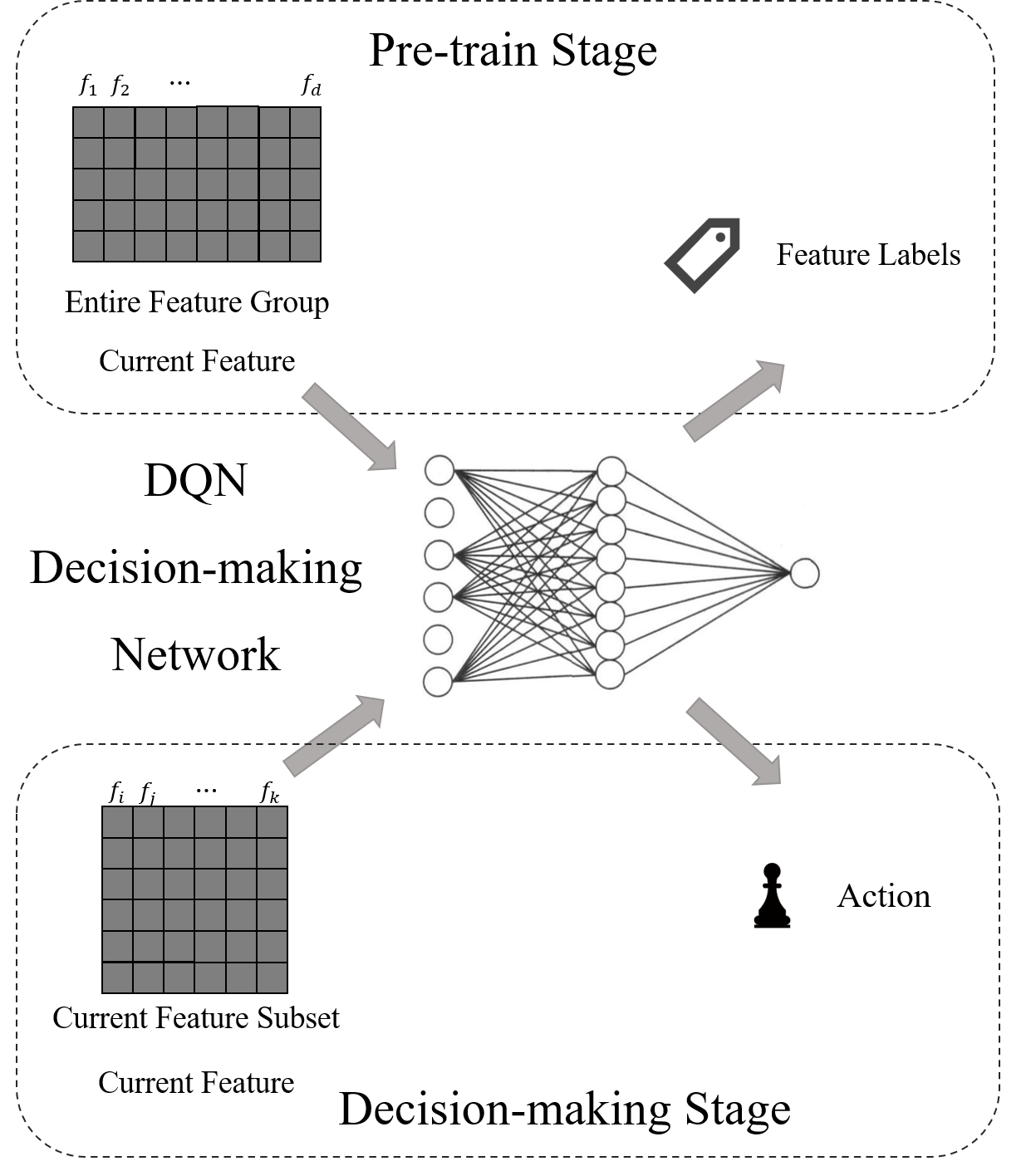}
\caption{The Structure of Pre-train Stage and Decision-making Stage.}
\label{fig:pretrain}
\end{figure}

The structure of the pre-train stage and decision-making stage can be shown in Figure~\ref{fig:pretrain}

From the perspective of forward propagation, these parameters of the decision network play the same role in both supervised learning and reinforcement learning frameworks. The labels in supervised learning also match the actual physical meaning involved in reinforcement learning feedback. So here parameters trained by supervised learning will also play an important role in reinforcement learning. For feature selection tasks, this injected information can help the agent avoid selecting features that are too far away from the pseudo feature set. In DQN, actions are generated by the decision-making network, and the action that can bring the maximum Q value in the current environment is selected, as shown in the following formula. The pre-training stage will adjust the network parameters ${\omega}_t$ in advance, thereby accelerating the network's adaptation to the training process.

\begin{equation}
  a_t={argmax}_aQ\left({s}_t,a;{\omega}_t\right)
\end{equation}

\subsubsection{$\varepsilon$-greedy Strategy Modification} \label{greedy}

At the same time, these feature labels are also used in the $\varepsilon$-greedy policy. The $\varepsilon$-greedy policy is a strategy used to balance exploration and exploitation when making decisions in a reinforcement learning environment. It allows the agent to learn through initial exploration, which is crucial for understanding the environment and constructing accurate Q-value functions. As the agent gains more experience and its Q-value estimates become more reliable, $\varepsilon$ can be gradually reduced over time to shift toward greater exploitation. The method we propose is to use these pseudo labels to replace part of the random selection, increase the balance in exploration and utilization, and inject external knowledge with a certain degree of reliability. The updated $\varepsilon$-greedy policy is expressed as follows:

\begin{equation}
\label{eq:greedy}
    \pi\left(\varepsilon\right)=\left\{
                \begin{array}{ll}
                  \ \ \ \ \ \ {\varepsilon}_1,\ \ \ \ \ \ \ \ \ \ \ \ \ \ \ \ \ \ P\left\{a_t={{A}}_i\mid i=0,1\right\}=\frac{1}{2}\\
                  \ \ \ \ \ \ {\varepsilon}_2,\ \ \ \ \ \ \ \ \ \ \ \ \ \ \ \ \ \ a_t={label}_{pesudo}\left(f_t\right)\\
                  1-{\varepsilon}_1-{\varepsilon}_2,\ \ \ \ \ \ \ \ \ \ \ a_t={max}_a{Q}^\ast\left({{s}}_t,a;\omega\right)
                \end{array}
              \right.
\end{equation}

\begin{figure}[htbp]
\centering
\includegraphics[width=0.5\linewidth]{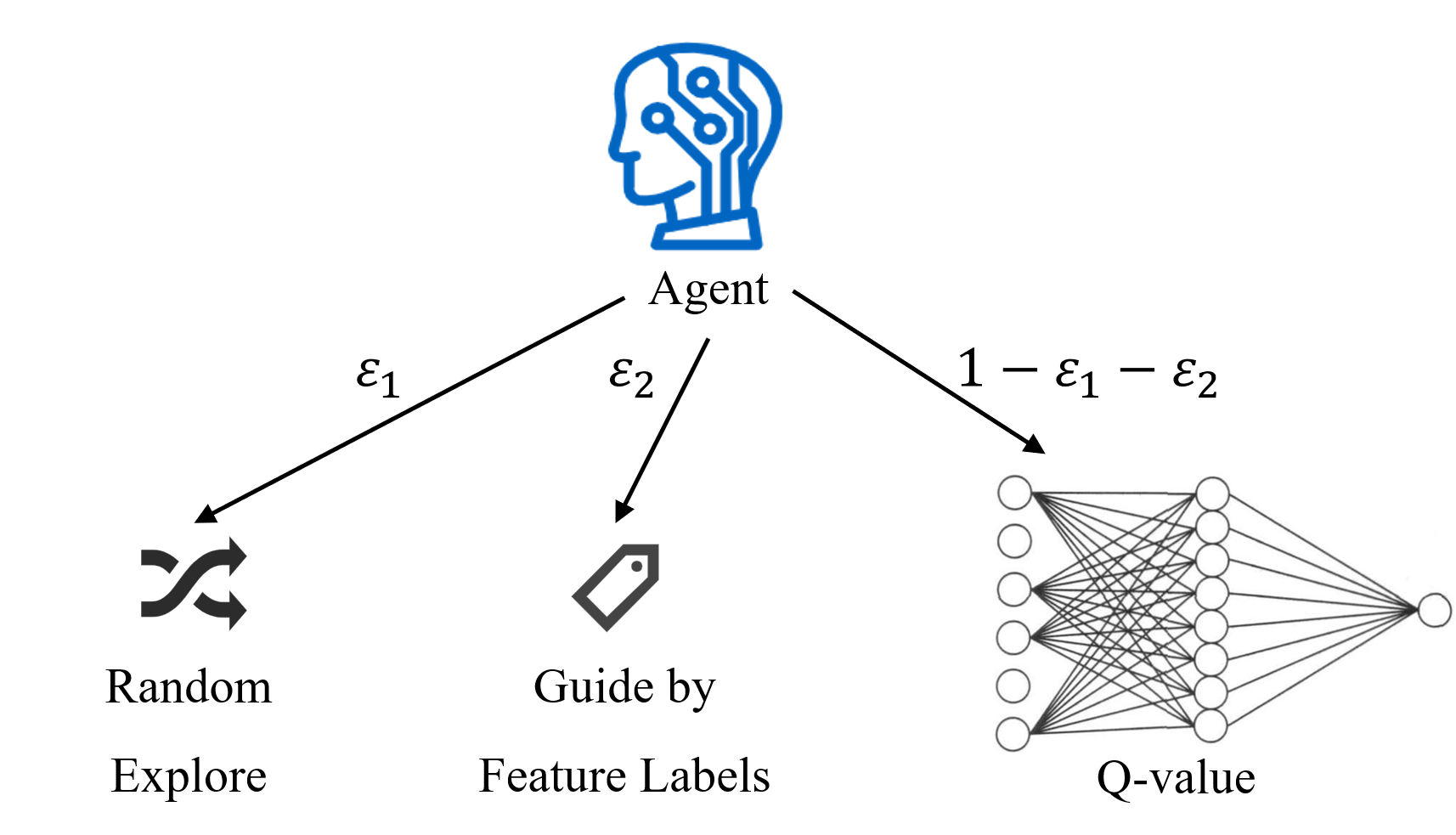}
\caption{The Structure of Renewed $\varepsilon$-greedy Policy.}
\label{fig:greedy}
\end{figure}

The structure of renewed $\varepsilon$-greedy policy can be shown in Figure~\ref{fig:greedy}

\subsection{Unsupervised Reward based on Pseudo In formation Injection and Matrix Reconstruction} \label{Unsupervised Reward}

The framework proposed in this article implements an unsupervised feature selection method that does not use the labels corresponding to the dataset for selection. This has two advantages: 1) There is no need to perform downstream tasks to generate a reward function after the agent makes a judgment, which can greatly save agent training time. 2) This feature selection method can be promoted on a large number of unlabeled data sets to improve the universality of the method. So here we use a new unsupervised reward function to answer the original reward function based on downstream task performance. Specifically, it is divided into two parts. The first part uses the matrix reconstruction method to compare the similarity between the existing selected feature subset and the entire feature set, while the other part compares the selection of the agent and the feature labels constructed by Knockoff for evaluation. The specific two parts are as follows.

\subsubsection{Matrix Reconstruction Reward} \label{ReconstructionReward}

The first part is the method of matrix reconstruction, which is shown as $R_{mr}\left({\bm{D}}_{sub}\right)$ in Equation ~\ref{reward}. Here we first train a convolutional Auto-Encoder using the existing selected feature subset and the entire feature set. Auto-encoders have been widely used in representation learning in an unsupervised manner~\cite{bengio2006greedy}, which take high-dimensional data as input and output low-dimensional representation vectors by minimizing the reconstruction loss. It can also solve the problem of changing feature set dimensions during the feature selection process. 

\begin{figure}[htbp]   
  \centering     
      \includegraphics[width=0.5\textwidth]{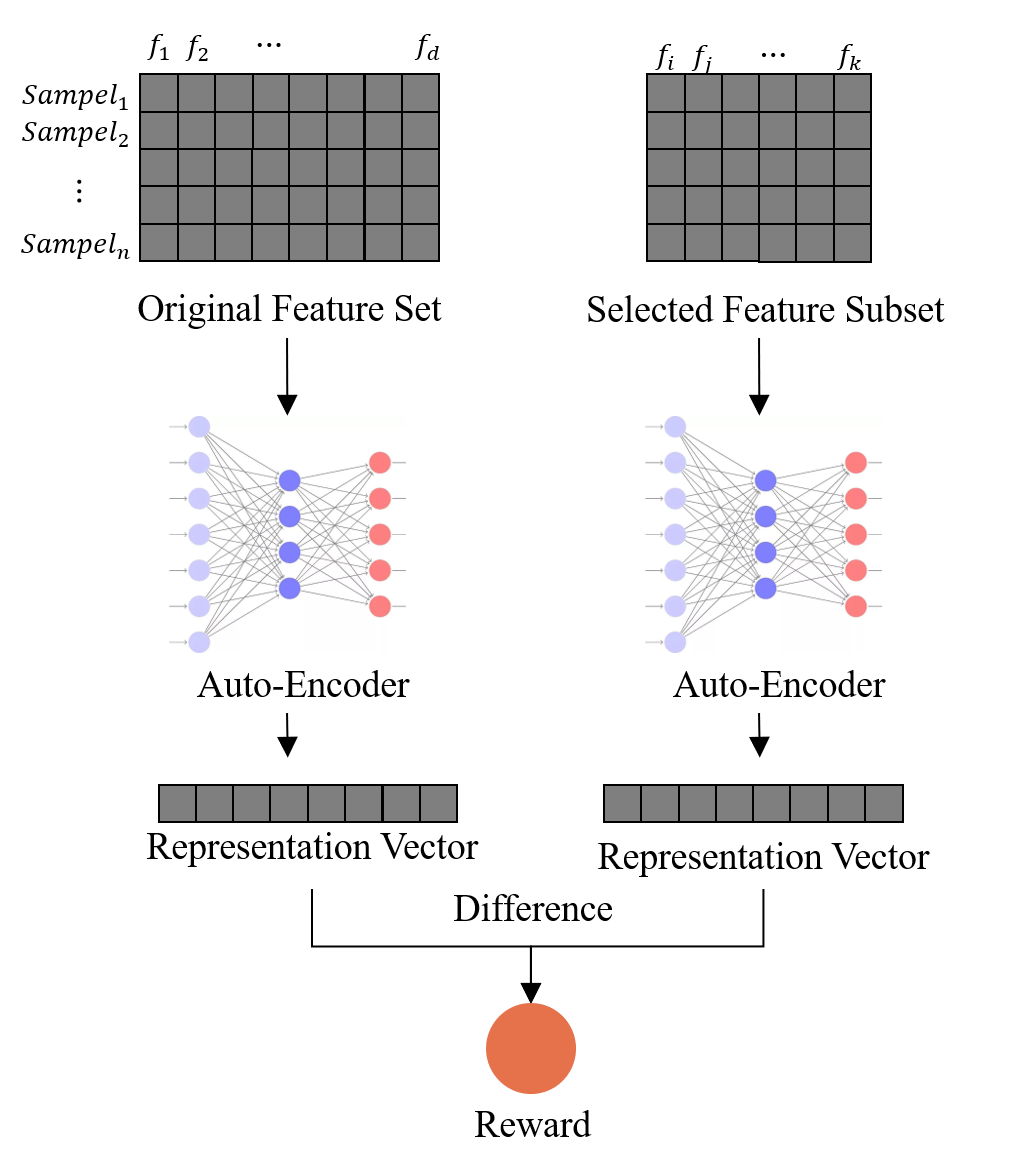}
  \caption{The Matrix Reconstruction Structure. The difference between two feature sets is used to construct the reward function, two Auto-Encoders are trained separately and then generate representation vectors of two feature sets respectively.}
  \label{reconstruction}
\end{figure}

Then the intermediate representation vectors of the two are extracted and compared, as shown in Figure~\ref{reconstruction}, which is the approach we have chosen. In addition, we also use the selected feature subset to directly reconstruct the original feature set and the loss will be used as a part of the reward function. However, the original feature set may contain some redundant features. Even if the selected feature subset contains all the information in the dataset, these redundant features may not be accurately restored. If this is the case, even if our feature subset achieves a specific purpose, that is, containing all information, it is possible to obtain a lower reward function, which is not conducive to the rapid convergence of reinforcement learning. We will discuss this in the experimental results.
If the selected feature subset can represent the entire feature set, the information contained in its representation vector will be very close to the entire feature representation vector. The specific expression is as follows:

\begin{equation}
  R_{mr}\left({D}_{sub}\right)=-\|AE\left({D}_{sub}\right),AE\left({D}\right)\|_2^2
\end{equation}

\subsubsection{Knockoff Feature Label Reward} \label{KnockoffReward}

In addition to the matrix reconstruction method, we introduce the feature labels constructed by Knockoff as a reference standard. Compared with the choice made by the agent, another part of the unsupervised reward is conducted, which is shown as $R_{pi}\left(f_i\right)$ in Equation ~\ref{reward}. The structure is shown in Figure~\ref{reward}.

\begin{figure}[htbp]   
  \centering     
      \includegraphics[width=0.5\textwidth]{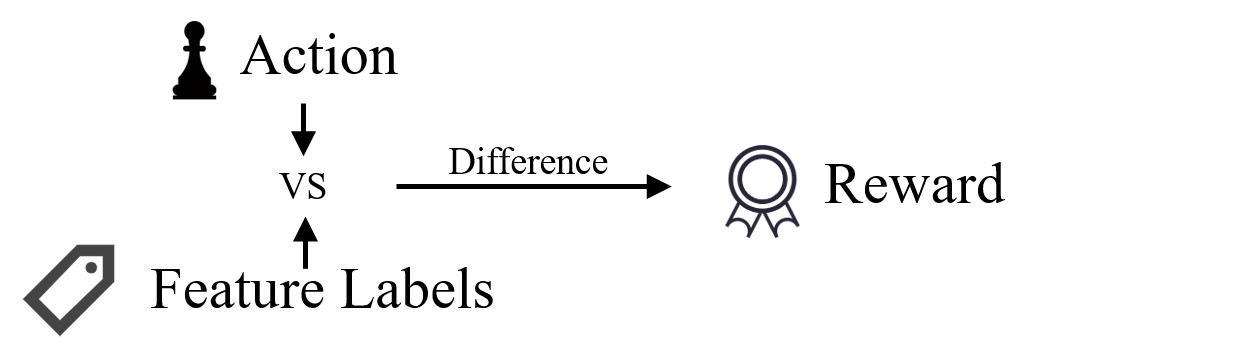}
  \caption{The Structure of Knockoff Feature Label Reward.}
  \label{reward}
\end{figure}

However, as mentioned above, we do not regard this label as solid information, that is, this label is not completely correct, so we have proposed two measures to ensure that the agent will not be affected by non-standard labels. 1) We only give negative incentives when the agent selects features with negative labels. As mentioned above, the negative label means that the feature is too close to the pseudo feature matrix, which means the agent should not select this feature. But when the agent does not select features with positive labels, we will not perform negative incentives, because features that are far away from the pseudo feature matrix are not necessarily suitable for selection. 2) We have added a decreasing mechanism. If the agent repeatedly selects features with negative labels, the negative incentives will gradually decrease. Because with the support of pre-training and negative incentives, the agent still insists on choosing this feature, indicating that this feature has merits, choosing this feature at this time should not be subject to severe negative incentives. This gives the entire reinforcement learning system the ability to correct errors. These two measures ensure the rationality of this part of the reward function. The specific expression is as follows:

\begin{equation}
  R_{pi}\left(f_i\right)=\tau^{time}\ast a_t\ast\left(1-{label}_{pi}\right)\ast p_{choose}
\end{equation}
Among them, $\tau$ represents the attenuation coefficient, $time$ represents the number of times the agent selects this feature, and $p_{choose}$ represents the probability of the agent selecting this feature.

\subsubsection{Features Redundancy Reward}
In Equation ~\ref{reward}, $R_{rd}(f_i)$ is employed to measure the redundancy within the feature subset, aiming to maintain a minimal level of redundant features. The calculation proceeds as follows:
\begin{equation}
  R_{rd}\left(f_i\right)=\frac{\sum_{k}\left|\rho\left(f_i,f_k\right)\right|}{d}
\end{equation}
where $\rho\left(f_i,f_k\right)$ denotes the correlation relationship.

\section{Experiments}

\subsection{Experimental Setup}
\subsubsection{Data Description}
To demonstrate the effectiveness of our framework, we conduct experiments using 26 datasets from UCIrvine~\cite{uci_dataset_2023}, CPLM~\cite{CPLM_2023}, Kaggle~\cite{howard_kaggle_2023}, and OpenML~\cite{Openml_dataset_2023}. 
These datasets are used for conducting regression and classification analyses. The corresponding statistics are presented in Table~\ref{tab:overallresult}.

\subsubsection{Evaluation Metrics}
Our task is to select an effective feature subset to achieve performance that is not inferior to or better than the entire feature set on the downstream tasks corresponding to the datasets. So here our test criterion for the selected feature subset is the indicator of the downstream task. 
For classification tasks, we utilize accuracy (ACC) as the key performance indicator. 
\begin{equation}
    ACC=\frac{TP+TN}{TP+TN+FP+FN}
\end{equation}
For regression tasks, we adopt the $\bm{\ell}_{2}$-norm as the metric of choice to evaluate the prediction error. 
\begin{equation}
    \bm{\ell}_{2} = \|\bm{y}_{real} - \bm{y}_{pred}\|_2^2
\end{equation}
To better evaluate the model performance among each dataset, we ranked each model on every dataset and calculated the Average Ranking.

\subsubsection{Baseline Algorithms}
We have selected the following methods as baselines for comparison with our model: (1) \textbf{LASSO}, a linear model suitable for high-dimensional data, implements feature selection and model simplification by adding L1 regularization penalties to the coefficients, encouraging some of them to shrink exactly to zero. (2) The \textbf{mRMR} method aims to select a feature set that is highly relevant to the target variable and minimally redundant among themselves. It optimizes the feature subset by balancing feature relevance and redundancy. (3) \textbf{GFS}, a feature selection method based on genetic algorithms, searches for the optimal feature subset by simulating natural evolution. This method is apt for dealing with complex and large-scale feature selection challenges. (4) \textbf{KBest} is a straightforward feature selection approach that chooses the best K features based on the statistical significance of each feature with the target variable. (5) \textbf{LASSONet} combines neural networks with LASSO for feature selection, using L1 regularization to identify and select features most crucial for model performance. (6) \textbf{RFE} selects features by progressively removing those contributing the least to the model. This recursive process eliminates one or more features in each iteration. (7) \textbf{MCDM}, a multi-criteria decision-making method, is employed in feature selection to balance and choose features under multiple evaluation criteria. (8) \textbf{MARLFS} utilizes a multi-agent reinforcement learning approach, identifying important features through the interaction and learning among agents. (9) \textbf{SADRLFS} adopts a single-agent reinforcement learning feature selection method, selecting features by combining the search order and extraction of states.

\subsubsection{Hyper parameters and Settings}
\begin{figure}[h]   
  \centering     
  \subfloat[German Credit Dataset]
  {
      \label{distance:GermanCredit}\includegraphics[width=0.5\textwidth]{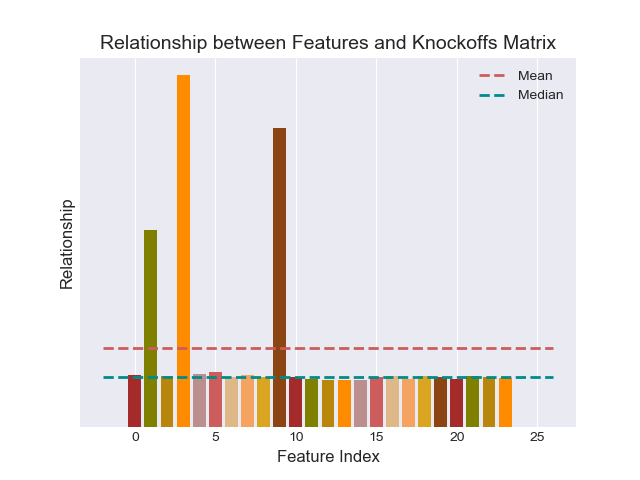}
  }
  \subfloat[Housing Boston Dataset]
  {
      \label{distance:HousingBoston}\includegraphics[width=0.5\textwidth]{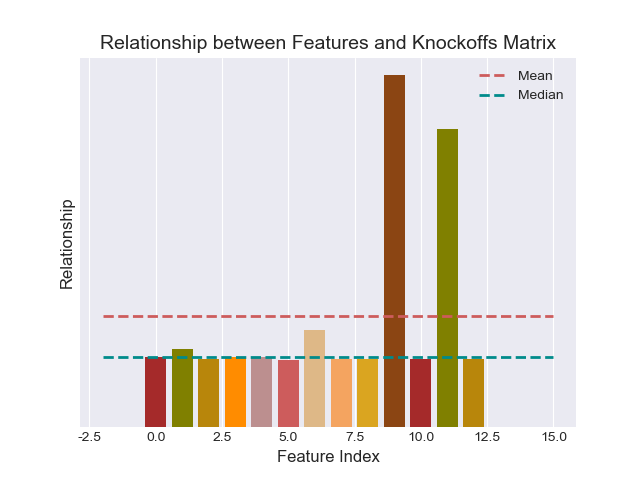}
  }
  \caption{The Distance Between Dataset Feature and Knockoff Matrix. Here we select 2 datasets as the example. German Credit~\ref{distance:GermanCredit} is a classification dataset and Housing Boston~\ref{distance:HousingBoston} is a regression dataset.}
  \label{distance}
\end{figure}

First, for the process of using the Knockoff method, we use a Gaussian sampler-based method to generate a pseudo matrix that is consistent with the original data size. We then label each feature based on its similarity to this matrix. Here, we consider features to be positive features whose distance to the Knockoff matrix is higher than the threshold. There are different ways to define this threshold, and we choose the median and the mean. However, if the median is used as the measurement threshold, there will be a situation where two features are close in distance but have completely opposite labels, as shown in Figure~\ref{distance}. These pictures are from 2 datasets we used for experiments.

As can be seen from the figures, the distance between only 2 or 3 features and the Knockoff matrix is significantly different from other features. If we use the median as the threshold (red line), the originally similar features will be divided into two categories, and the originally different features will be classified into one category. This will cause label confusion. If incorrect information is used for the next step, an incorrect feature selection will result.
Regarding the process of introducing feature labels into the greedy strategy, at the beginning we selected 5\% of the cases for random exploration, 5\% of the cases gave results according to the feature labels, and the remaining 90\% of the cases were used for Q network selection. The probability of random exploration and label selection will decrease as training progresses. There is a study on $\varepsilon$-greedy policy parameters in Section~\ref{greedyParameters}.
For the matrix reconstruction process, an auto-encoder is first trained by the selected feature subset and the original feature set separately, and then MSE is used to calculate the difference between the two representation vectors as part of the reward function. 
For the reinforcement learning training process, we explore each dataset ten times, that is, the number of training steps is ten times the number of features. After each complete exploration of the dataset, the corresponding downstream tasks are used to evaluate the currently selected feature subset.

\subsubsection{Experimental Environment}
All experiments were conducted on the Ubuntu 22.04.3 LTS operating system, 13th Gen Intel(R) Core(TM) i9-13900KF CPU, with the framework of Python 3.11.5 and PyTorch 2.0.1.

\subsection{Experimental Results}

\begin{table}[h]
  \begin{center}
    \caption{Experiment Results Compared to Supervised Feature Selection Method.}
    \label{tab:overallresult}
        \resizebox{\textwidth}{!}{\begin{tabular}{ccccccccccccccc}
            \toprule
            Dataset & Source & Task & Samples & Features & LASSO & mRMR & GFS & KBest & LASSONet & RFE & MCDM & MARLFS & SADRLFS & \textbf{KGPTFS} \\
            \midrule
            Carto & Kaggle & C & 15120 & 54 & 56.75\% & 86.64\% & 79.03\% & 85.52\% & 49.07\% & 86.84\% & 79.89\% & \textbf{88.76\%} & 88.56\% & 88.10\%\\
            Amazon Employee & Kaggle & C & 32769 & 9 & 93.44\% & 91.70\% & 93.56\% & 93.96\% & 93.83\% & 92.65\% & 93.74\% & 94.90\% & \textbf{95.48\%} & 95.03\% \\
            Glycation & CPLM & C & 630 & 402 & 79.37\% & 80.95\% & 79.37\% & 82.54\% & 80.95\% & 77.78\% & 85.71\% & 82.54\% & \textbf{87.30\%} & 85.71\% \\
            SpectF & UCIrvine & C & 267 & 44 & 81.48\% & 77.78\% & 88.89\% & 85.16\% & 85.16\% & 88.89\% & 81.48\% & 88.89\% & \textbf{92.75\%} & 92.59\% \\
            German Credit & UCIrvine & C & 1000 & 24 & 77.00\% & 79.00\% & 72.00\% & 76.00\% & 70.00\% & 74.00\% & 75.00\% & 81.00\% & 80.00\% & \textbf{82.00\%} \\
            UCI Credit & UCIrvine & C & 30000 & 23 & 78.97\% & 81.10\% & 81.53\% & 81.63\% & 79.27\% & 81.37\% & 80.20\% & 82.57\% & 82.10\% & \textbf{82.70\%} \\
            Spam Base & UCIrvine & C & 4601 & 57 & 93.06\% & 94.36\% & 93.93\% & 95.44\% & 93.06\% & 95.88\% & 95.01\% & 97.18\% & 95.66\% & \textbf{97.61\%} \\
            Ionosphere & UCIrvine & C & 351 & 34 & 91.67\% & 88.89\% & 86.11\% & 91.67\% & 83.33\% & 91.67\% & 88.89\% & 91.67\% & 91.67\% & \textbf{94.44\%} \\
            Human Activity & UCIrvine & C & 10299 & 562 & 97.18\% & 97.67\% & 98.06\% & 97.86\% & 97.38\% & 96.70\% & 97.38\% & 98.15\% & 98.16\% & \textbf{98.44\%} \\
            Higgs Boson & UCIrvine & C & 50000 & 28 & 67.86\% & 64.58\% & 64.92\% & 62.94\% & 69.50\% & 70.14\% & 63.08\% & 70.76\% & 72.28\% & \textbf{72.34\%} \\
            PimaIndian & Kaggle & C & 768 & 8 & 72.73\% & 76.62\% & 68.83\% & 71.43\% & 70.13\% & 75.32\% & 75.32\% & 75.32\% & 77.54\% & \textbf{77.92\%} \\
            Messidor Feature & UCIrvine & C & 1151 & 19 & 61.21\% & 65.52\% & 62.93\% & 59.48\% & 58.62\% & 67.24\% & 66.38\% & \textbf{72.41\%} & 69.83\% & 70.69\% \\
            Wine Quality Red & UCIrvine & C & 999 & 11 & 69.00\% & 77.00\% & 57.00\% & 76.00\% & 75.00\% & 78.00\% & 80.00\% & 82.00\% & 77.00\% & \textbf{83.00\%} \\
            Wine Quality White & UCIrvine & C & 4898 & 11 & 68.57\% & 65.10\% & 49.18\% & 68.37\% & 65.31\% & 67.76\% & 68.16\% & 70.61\% & \textbf{72.24\%} & 71.02\% \\
            yeast & UCIrvine & C & 1484 & 8 & 84.56\% & 87.25\% & 85.23\% & 85.91\% & 83.89\% & 87.25\% & 85.23\% & 87.92\% & 90.60\% & \textbf{91.28\%} \\
            phpDYCOet & OpenML & C & 34465 & 118 & 96.75\% & 96.72\% & 96.23\% & 95.27\% & 95.62\% & 96.67\% & 96.14\% & 97.07\% & 96.98\% & \textbf{97.39\%} \\
            \midrule
            Housing California & UCIrvine & R & 20640 & 8 & 0.7629 & 0.7867 & 0.6962 & 0.7503 & 0.9971 & 0.7365 & 0.7647 & \textbf{0.5917} & 0.6104 & 0.6027 \\
            Housing Boston & Kaggle & R & 506 & 13 & 21.3557 & 16.9978 & 16.5129 & 15.9910 & 34.8086 & 12.8979 & 15.7711 & 11.6156 & 11.0865 & \textbf{9.5343} \\
            Airfoil & UCIrvine & R & 1503 & 5 & 47.8969 & 38.9785 & 13.6322 & 12.7275 & 38.6121 & 20.6711 & 15.9771 & 14.6501 & 4.6571 & \textbf{4.3359} \\
            Openml 618 & OpenML & R & 1000 & 50 & 0.2809 & 0.2098 & 0.2050 & 0.2158 & 1.0071 & 0.2490 & 0.1901 & 0.1950 & 0.2208 & \textbf{0.1822} \\
            Openml 589 & OpenML & R & 1000 & 25 & 0.1642 & 0.1512 & 0.1724 & 0.1721 & 0.3675 & 0.1691 & 0.1778 & 0.1455 & 0.1762 & \textbf{0.1429} \\
            Openml 616 & OpenML & R & 500 & 50 & 0.2888 & 0.2483 & 0.2176 & 0.2091 & 0.3368 & 0.2314 & 0.2520 & 0.1884 & 0.1809 & \textbf{0.1778} \\
            Openml 607 & OpenML & R & 1000 & 50 & 0.2088 & 0.2326 & 0.2189 & 0.2328 & 1.3336 & 0.2034 & 0.1797 & 0.1796 & 0.1865 & \textbf{0.1682} \\
            Openml 620 & OpenML & R & 1000 & 25 & 0.3008 & 03534. & 0.1718 & 0.1551 & 0.1721 & 0.1661 & 0.1591 & \textbf{0.1504} & 0.1957 & 0.1518 \\
            Openml 637 & OpenML & R & 500 & 50 & 0.2726 & 0.5098 & 0.2631 & 0.2553 & 0.2539 & 0.2624 & 0.2322 & 0.2394 & 0.2281 & \textbf{0.2232} \\
            Openml 586 & OpenML & R & 1000 & 25 & 0.2408 & 0.2596 & 0.2393 & 0.2409 & 0.2371 & 0.2526 & 0.2350 & 0.2261 & \textbf{0.1634} & 0.2215 \\
            \midrule
            \textbf{Average Ranking} & - & - & - & - & 7.73 & 6.96 & 6.85 & 5.96 & 7.92 & 6.12 & 6.04 & 2.88 & 3.12 & \textbf{1.42} \\
            \bottomrule
        \end{tabular}}
    \end{center}
\end{table}

\subsubsection{Overall Comparison}
This experiment aims to answer: {\itshape Can our method effectively select a quality feature subset and improve a downstream task?}
We compare the KGPTFS method with several baseline supervised feature selection methods. The experiment involves a comprehensive assessment over 26 diverse datasets, examining the efficacy of various feature transformation methods.
The results are detailed in Table~\ref{tab:overallresult}, which shows the performance of the unsupervised KGPTFS method compared to traditional supervised methods across different datasets. The table highlights the effectiveness of the selected features by our method in improving the performance of downstream tasks. The average ranking shows that the KGPTFS method out-performance in majority of datasets.
Our findings indicate that the features selected by KGPTFS are not only comparable in performance to those chosen by supervised methods but also offer advantages in terms of time efficiency and broader applicability. Unlike supervised methods, KGPTFS does not require downstream tasks for verification, saving significant time, especially with large datasets. Furthermore, the unsupervised nature of KGPTFS allows its application to unlabeled datasets, greatly expanding its usability in real-world scenarios where manual labeling is time-consuming, and unlabeled data sets are common.
The experiment validates the effectiveness of the KGPTFS method in feature selection for downstream tasks. Its unsupervised approach provides notable benefits in terms of time efficiency and applicability to a wide range of datasets, particularly in engineering practices where unlabeled data are prevalent. The method's independence from downstream task requirements represents a significant advancement over traditional, supervised feature selection methods.

\begin{figure}[!h]
\centering
\includegraphics[width=0.8\textwidth]{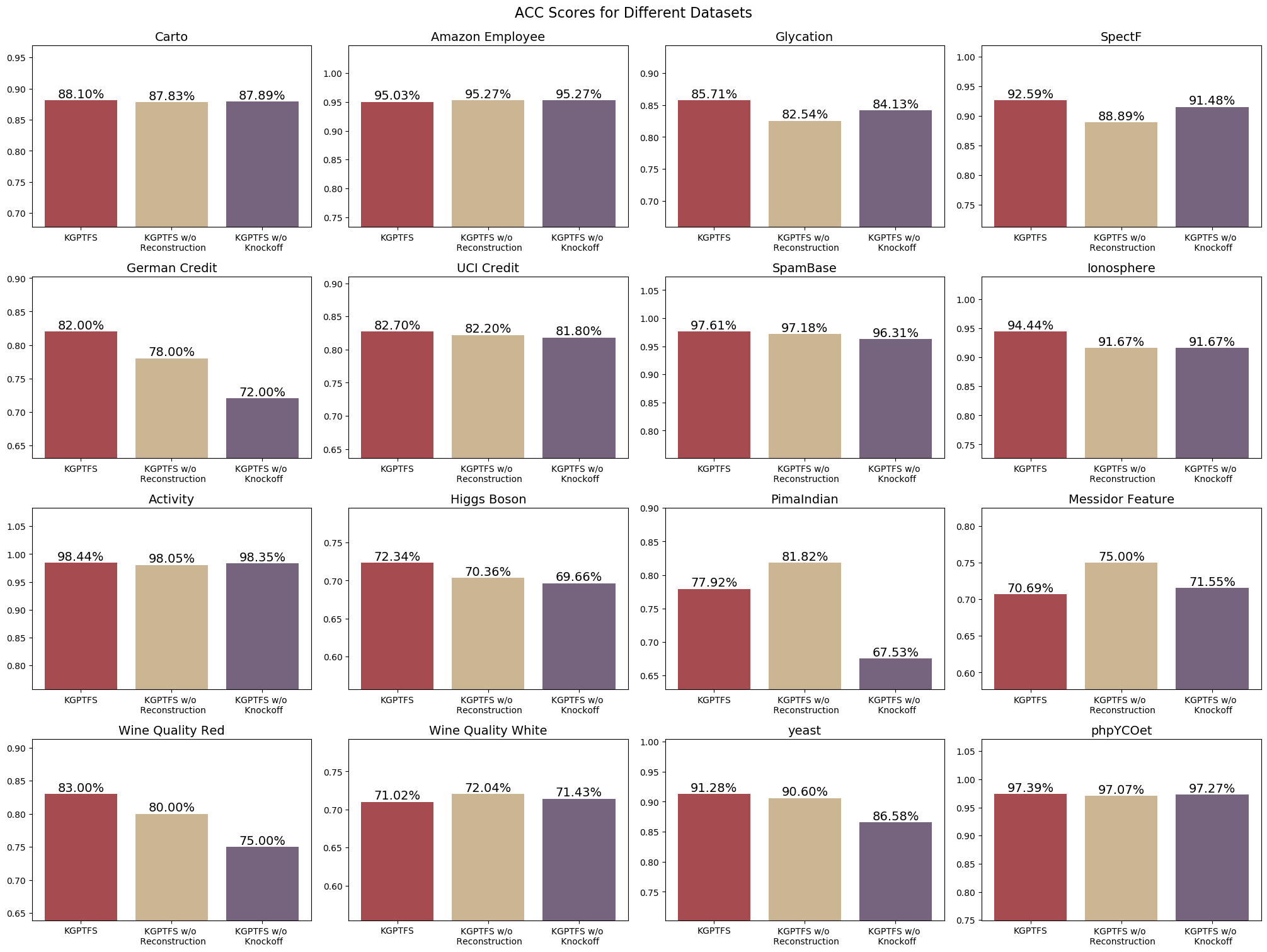}
\caption{Experiment Results of Study on Unsupervised Reward (Classification Task).}
\label{fig:AblationACC}
\end{figure}

\begin{figure}[!h]
\centering
\includegraphics[width=0.8\textwidth]{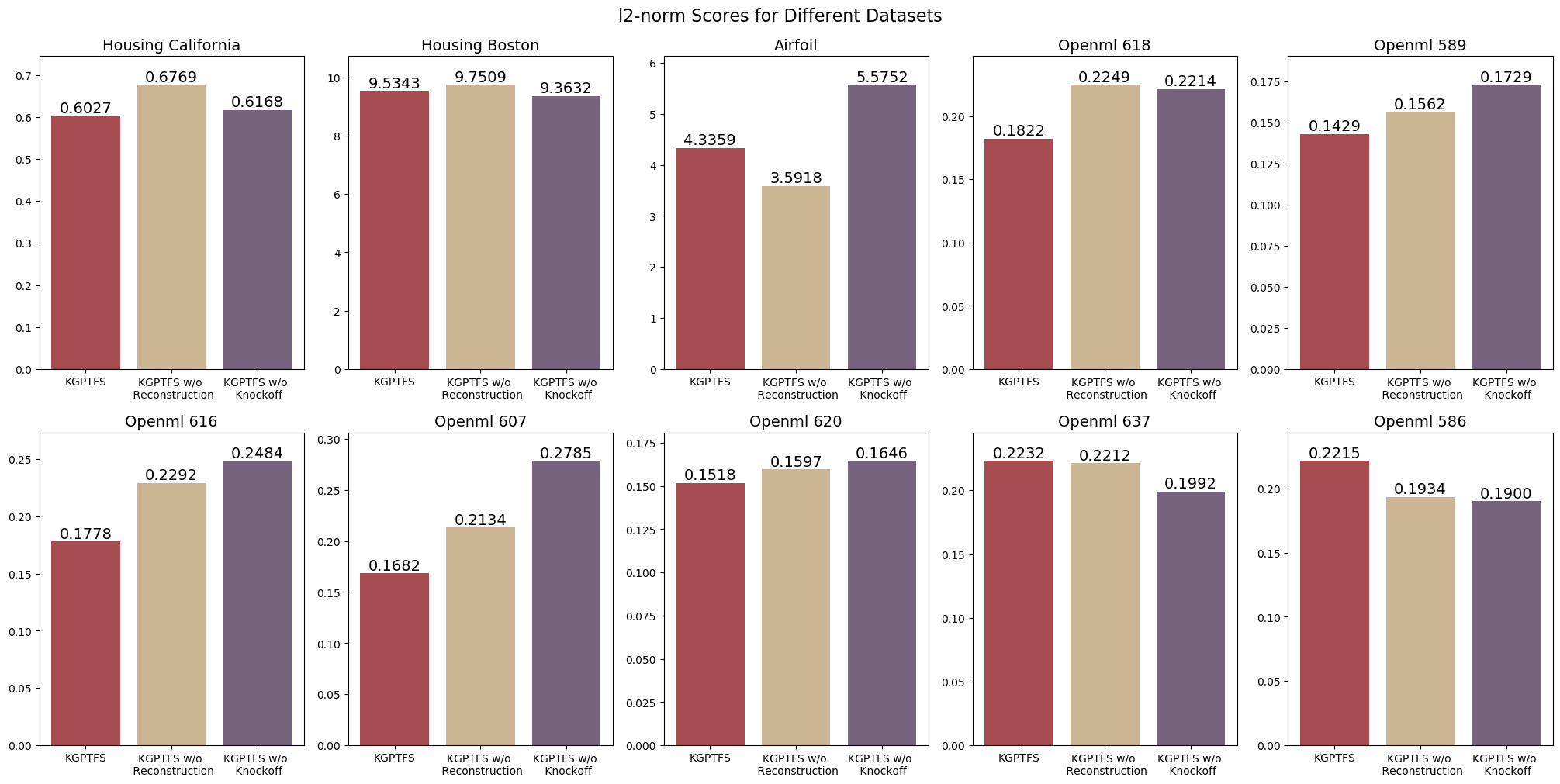}
\caption{Experiment Results of Study on Unsupervised Reward (Regression Task).}
\label{fig:Ablationnorm}
\end{figure}

\subsubsection{Ablation Study}
This experiment aims to answer: {\itshape How essential are each component's contributions to our model's performance?}
What's more, we conducted ablation experiments to validate the effectiveness of each component of our model, and the specific results are presented in Figure~\ref{fig:UnsupervisedRewardACC} and Figure~\ref{fig:UnsupervisedRewardnorm}. The experimental results indicate that our method consistently outperforms others on the majority of datasets. This suggests that, in most cases, the combination of knockoff pseudo information injection and matrix reconstruction is indispensable.

\subsubsection{Study on Unsupervised Reward}
This experiment aims to answer: {\itshape What is the impact of different unsupervised reward strategies on our feature selection method's effectiveness?}

\begin{figure}[h]
\centering
\includegraphics[width=0.8\textwidth]{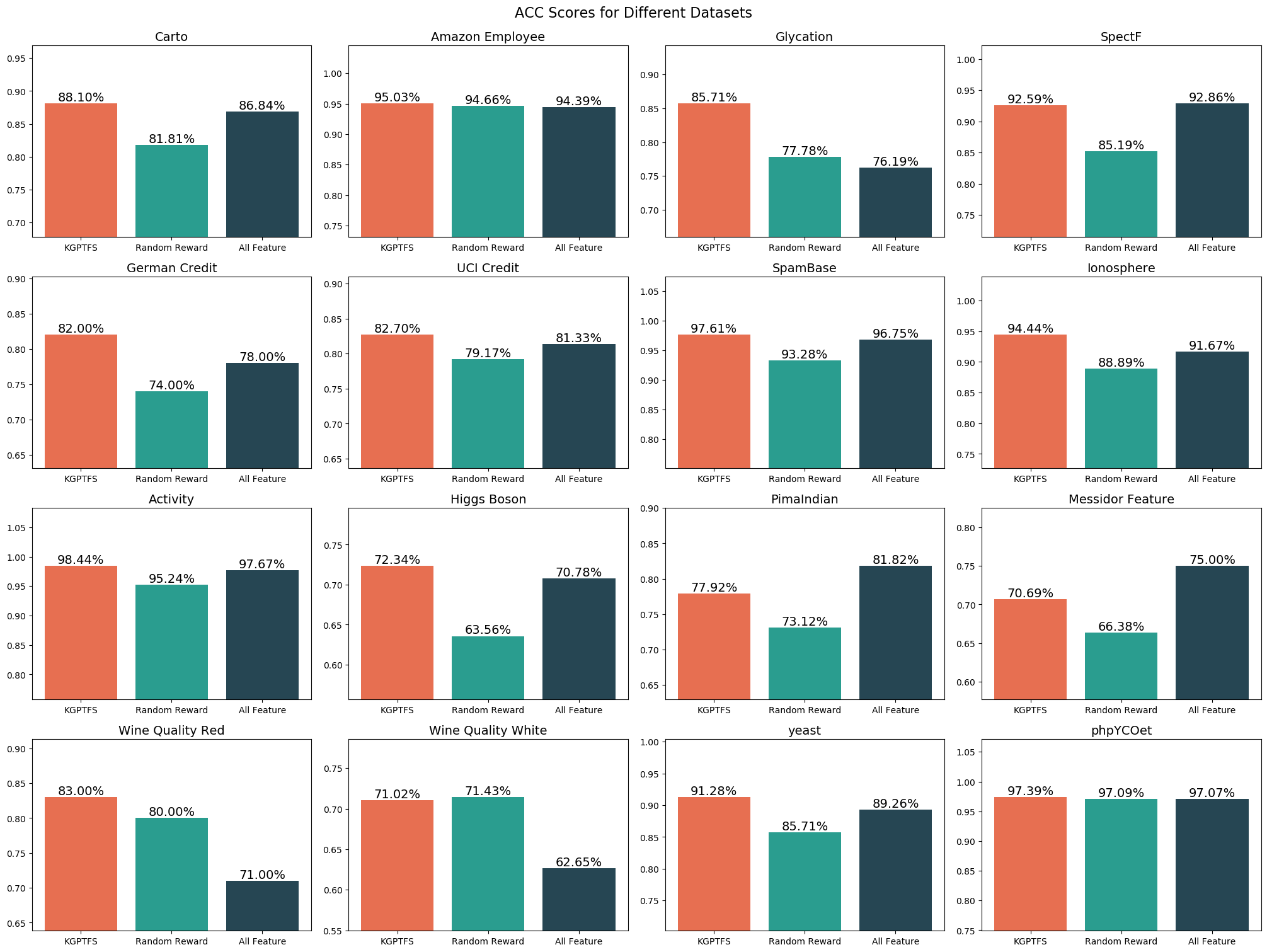}
\caption{Experiment Results of Study on Unsupervised Reward (Classification Task).}
\label{fig:UnsupervisedRewardACC}
\end{figure}

\begin{figure}[h]
\centering
\includegraphics[width=0.8\textwidth]{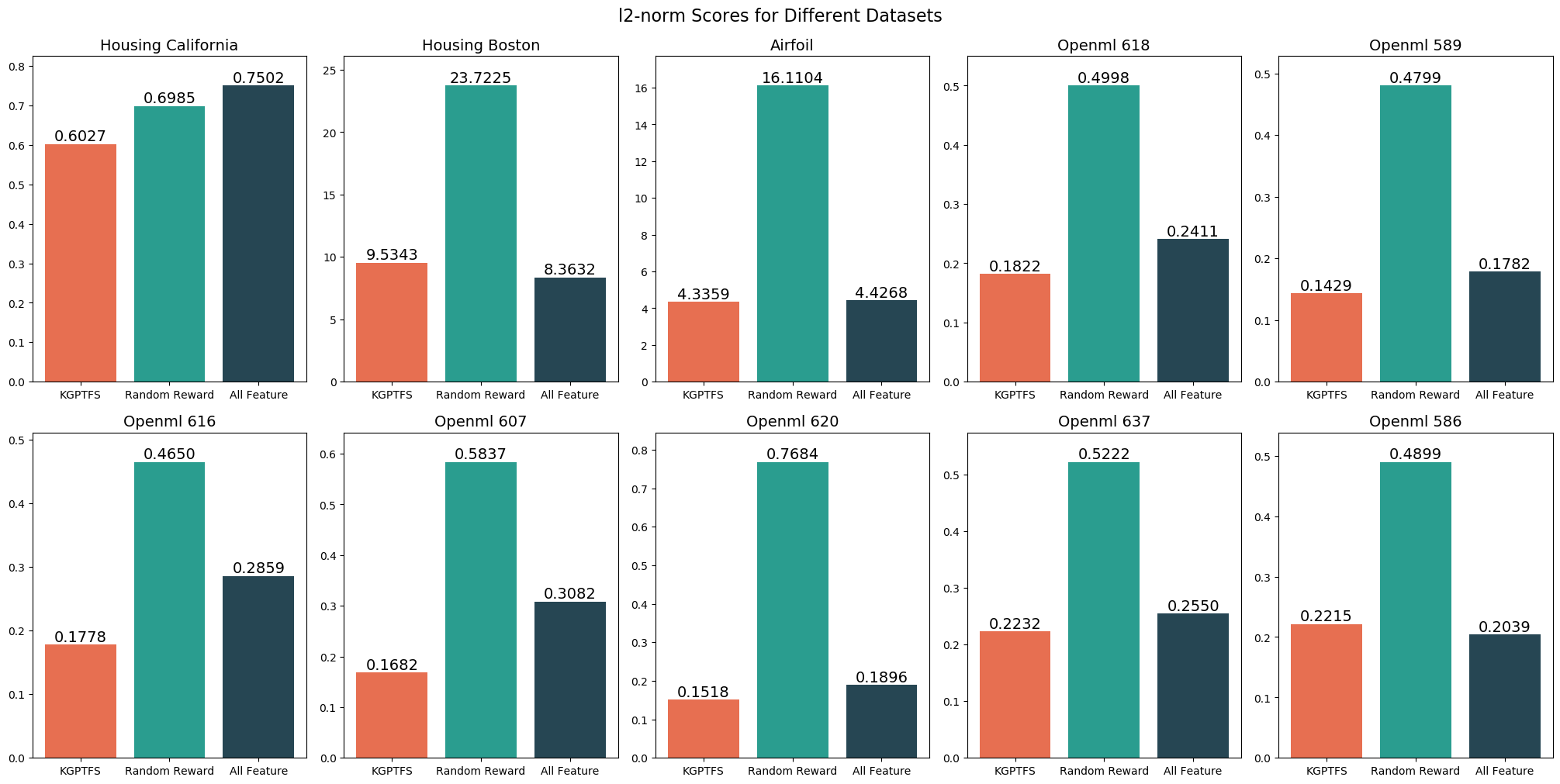}
\caption{Experiment Results of Study on Unsupervised Reward (Regression Task).}
\label{fig:UnsupervisedRewardnorm}
\end{figure}

To verify the necessity of the proposed unsupervised method, we conduct ablation experiments on unsupervised reward. In this experiment, our comparison model is the proposed feature selection method, and then three different methods are designed for comparison. The first is to use random rewards for the agent. This experiment is mainly to verify that the unsupervised reward function we proposed is meaningful, intrinsically related to the features, and is not a random reward function. Here, the agent will be randomly given a value between -1 and 1 as a reward. Secondly, we do not use reinforcement learning methods for feature selection but randomly select 50\% of the features for downstream tasks to verify the effectiveness of the automated feature selection process. Finally, we do not perform the feature selection process but use the entire feature set for downstream tasks as a baseline model for comparison.

Our experimental results are shown in Figure~\ref{fig:UnsupervisedRewardACC} and Figure~\ref{fig:UnsupervisedRewardnorm}. It can be seen from this that the feature subset selected by the agent using the random reward function does not perform satisfactorily on downstream tasks, so it shows that our unsupervised reward function is intrinsically related to the features and is meaningful. After random feature selection, the performance of this subset of features on downstream tasks will be even worse, so it is very necessary to use reinforcement learning for automated feature selection. When using all features for downstream tasks, we found that the performance of feature selection using unsupervised methods is similar to that of all features, which shows that our method selects truly useful features and removes redundant features, completing the task perfectly. 

\subsubsection{Study on Pseudo Information Component Importance}
\begin{figure}[htbp]
\centering
\includegraphics[width=0.8\textwidth]{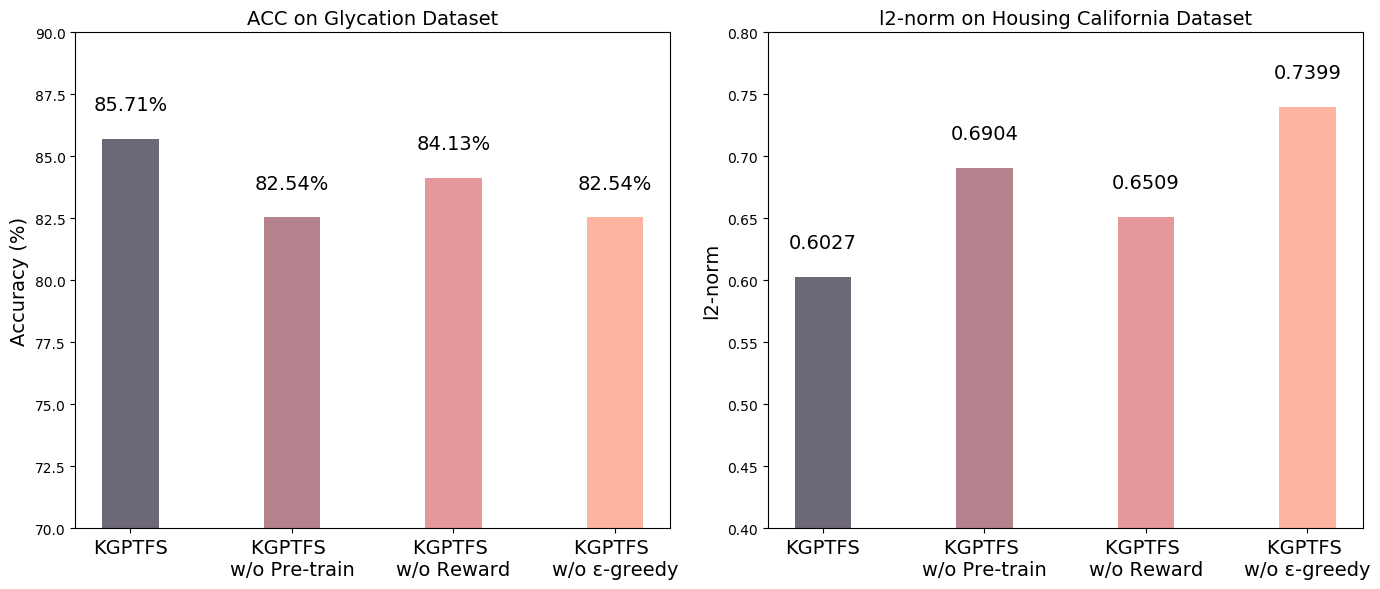}
\caption{Experiment Results of Study on Pseudo Information Component Importance.}
\label{fig:PseudoInformationComponentImportance}
\vspace{-0.3cm}
\end{figure}

This experiment aims to answer: {\itshape How does the exclusion of each module in the Knockoff pseudo-information injection process affect the overall experimental results?}
This experiment mainly focuses on the Knockoff pseudo-information injection part. There are 3 modules in total, which are: not including the use of feature labels for pre-training, not including the use of the Knockoff reward function, and not including the use of Knockoff pseudo-information to guide the greedy strategy. The obtained results are shown in Figure~\ref{fig:PseudoInformationComponentImportance}. The results clearly demonstrate the critical importance of all three methods that employ pseudo information. This significance arises because each of these three components integrates information from feature labels provided by the knockoff process. This information indirectly assists the agent in decision-making. Moreover, the infusion of this information into various segments of the reinforcement learning process enhances the robustness of the results, ensuring a more comprehensive and effective information injection.

\subsubsection{Study on $\varepsilon$-greedy Policy Parameters} \label{greedyParameters}
This experiment aims to answer: {\itshape How does varying the parameters of the $\varepsilon$-greedy strategy affect the performance of our framework }

\begin{figure}[htbp]
\centering
\includegraphics[width=0.8\textwidth]{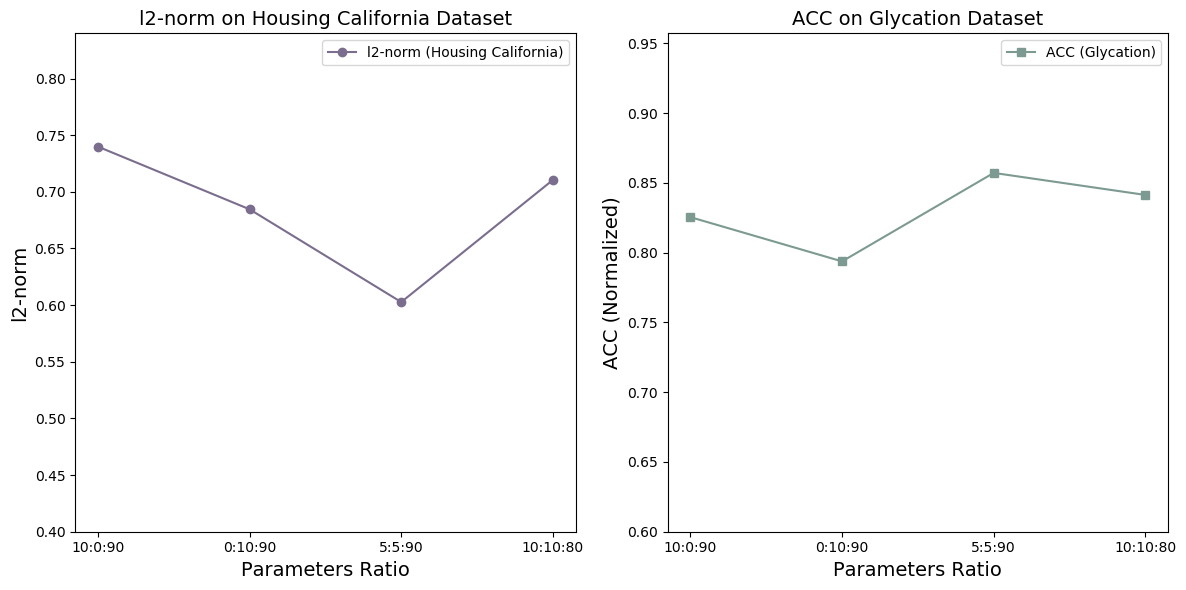}
\caption{$\varepsilon$-greedy Policy Parameters Experiment Results.}
\label{fig:greedyparameters}
\end{figure}

In this section, we examine the implementation of the $\varepsilon$-greedy strategy. We selected various parameter combinations to test our framework. These three ratios correspond to the three parameters described in Equation~\ref{eq:greedy} and Figure~\ref{fig:greedy}, with four distinct combinations in total. The 10:0:90 combination indicates that no knockoff information was utilized during training, whereas the 0:10:90 combination indicates the absence of random selection.
The results obtained are displayed in Figure~\ref{fig:greedyparameters}. The two sub-figures represent the outcomes on different datasets. For sub-figure 1, the Housing California dataset corresponds to a regression task with the $\bm{\ell}_{2}$-norm as the evaluation metric (the lower, the better), indicating that the combination of 5:5:90 yields the best performance. Similarly, for sub-figure 2, the Glycation dataset pertains to a classification task (with the evaluation metric ACC where higher is better), and once again, the 5:5:90 combination shows the best results. Therefore, we believe that appropriately substituting random selection with knockoff-based feature labeling in the $\varepsilon$-greedy policy offers certain advantages.

\subsubsection{Study on Different Matrix Reconstruction Method}
This experiment aims to answer: {\itshape What is the effectiveness and necessity of the matrix reconstruction component?}
This is followed by the second part, which mainly focuses on the matrix reconstruction part. In Chapter~\ref{Methodology}, we mentioned that we use comparison to generate the reward function because this can more accurately compare the information difference between the feature subset and the original feature set. Here, we use another way to perform reward function construction, which is to use feature subsets to reconstruct the original feature set and use the Auto-Encoder's loss function as part of the reward function. At the same time, to verify the necessity of this unsupervised reward function, we eliminated it and conducted ablation experiments.

\begin{figure}[htbp]
\centering
\includegraphics[width=0.8\textwidth]{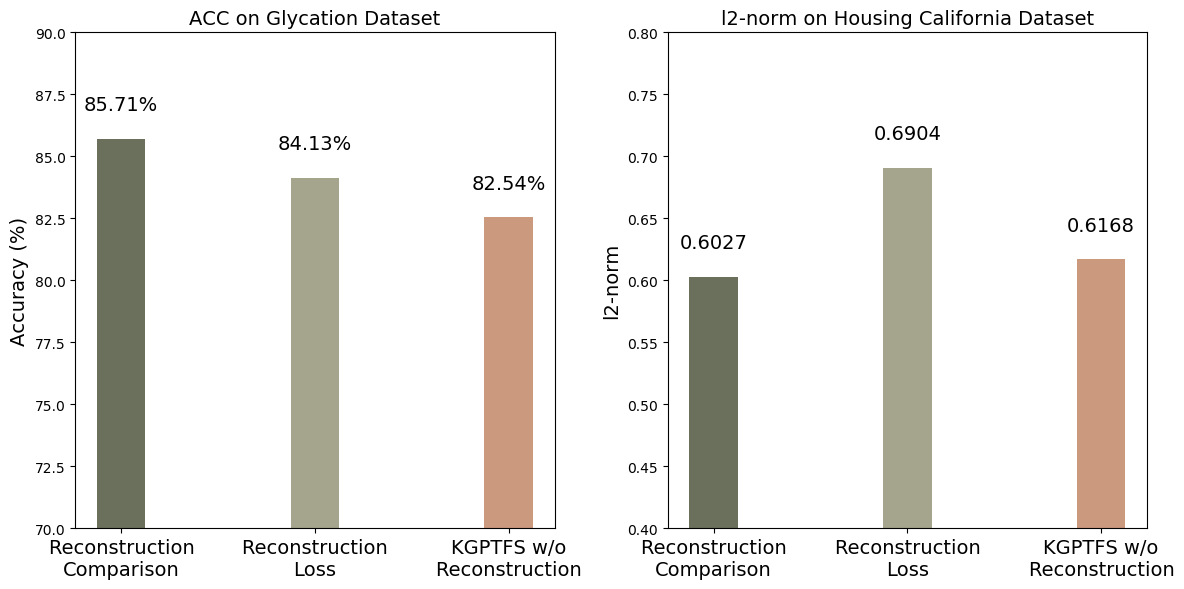}
\caption{Experiment Results of Different Matrix Reconstruction Method.}
\label{fig:DifferentMatrixReconstruction}
\end{figure}

The obtained results are shown in Figure~\ref{fig:DifferentMatrixReconstruction}. We can see that the feature subset selected by using the auto-encoder loss function has lower performance on downstream tasks. This is because the feature subset is reconstructed during reconstruction. Two conversions are performed, but the original feature set is not converted, so this method will increase the difference between the two sets, which will have a bad impact. At the same time, even if the feature subset contains most of the useful information, the redundant features in the original feature set cannot be completely restored based on this part of the information. Because redundant features have no rules. If the redundant features cannot be completely restored, the loss function of the auto-encoder will increase, but this is not what we expect. Failure to restore redundant features should not be punished in feature selection. Our experimental results also verify this view. Finally, after eliminating the matrix reconstruction module, the performance decline on downstream tasks also illustrates the necessity of this part.

\subsubsection{Study on Generalization}
\begin{figure}[htbp]
\centering
\includegraphics[width=0.8\textwidth]{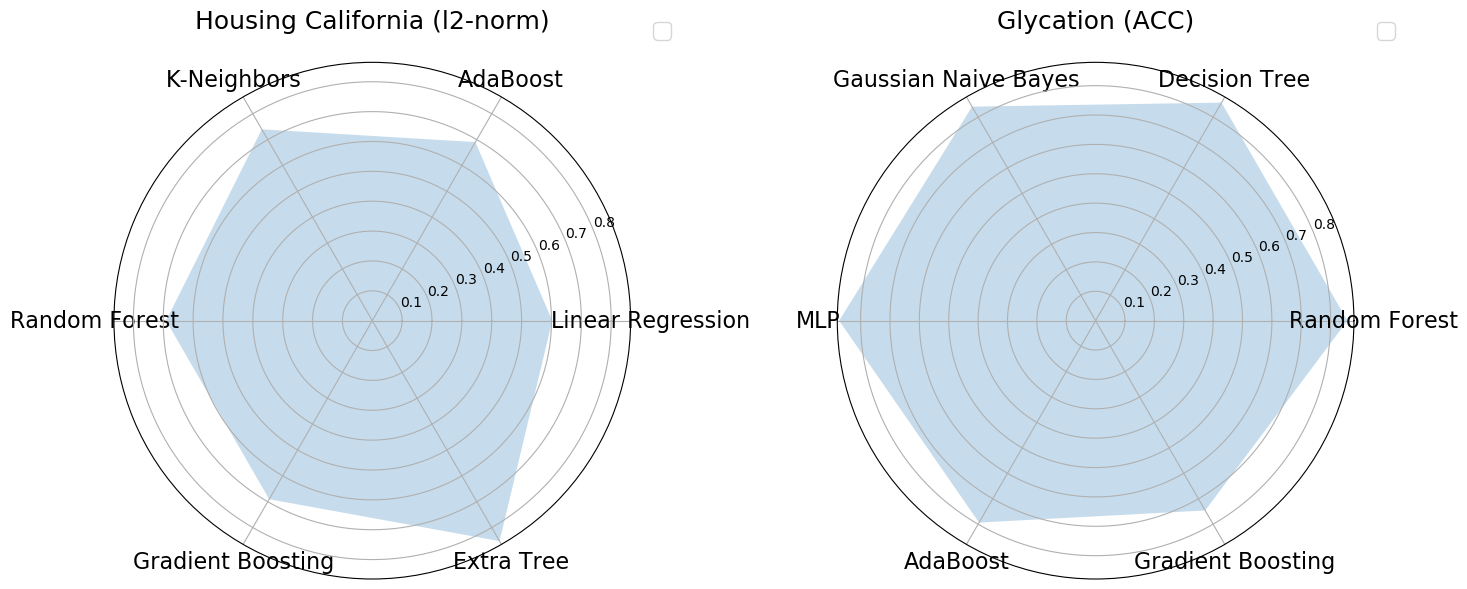}
\caption{Generalization Experiment Results.}
\label{fig:Generalization}
\end{figure}

This experiment aims to answer: {\itshape Can our method fairly performance on different classification/regression methods?}
We conduct a generalization experiment to verify the universality of our proposed feature selection method. Our automated unsupervised feature selection method does not depend on downstream tasks, so it is also a model-free structure. To verify the universality of this method, we used different classification methods to verify the selected feature subsets. What we have chosen are simple machine learning methods.
The final results are reflected in the Figure~\ref{fig:Generalization}. For classification tasks and regression tasks, we used 6 models for each experiment. We can see from the results that the selected features themselves perform similarly on different models, and the final downstream task results are not much different. Therefore, our automated unsupervised method is highly practical, which also reflects the broad scope of application of our method.

\subsubsection{Study on Running Time}
This experiment aims to answer: {\itshape Can our method save time compared to the supervised feature selection method?}

\begin{figure}[htbp]
\centering
\includegraphics[width=0.7\textwidth]{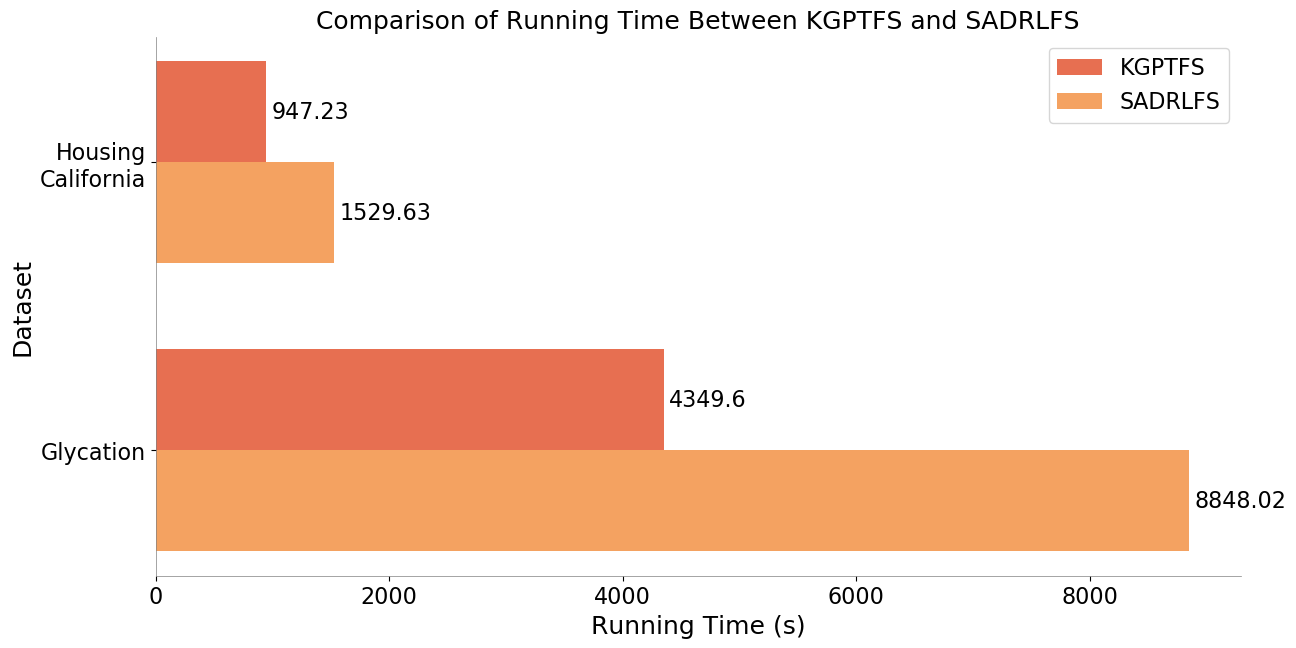}
\caption{Experiment Results of Running Time.}
\label{fig:RunningTime}
\end{figure}

To demonstrate the efficiency and time-saving capabilities of our framework, we conducted a comparative analysis of the running times between our unsupervised method and conventional supervised methods. The results of this analysis are comprehensively presented in Figure~\ref{fig:RunningTime}.
The results clearly indicates that our unsupervised approach operates at a faster pace compared to its supervised counterpart. This enhanced speed can be attributed primarily to the elimination of downstream tasks following each selection step in the unsupervised method. This aspect not only streamlines the feature selection process but also reduces the overall computational burden.
Furthermore, the time efficiency of our unsupervised method offers substantial benefits in practical applications, especially in scenarios where rapid data processing and decision-making are crucial. By bypassing the need for repetitive verification processes inherent in supervised methods, our approach facilitates quicker adaptation to new datasets and faster iteration cycles. This advantage makes our method particularly valuable in dynamic environments or when dealing with large datasets, where time is a critical resource.

\section{Related Works}

In the realm of feature selection, various methods can be broadly categorized into supervised and unsupervised approaches.
Similarity-based methods assess features by gauging their similarity to class labels or among themselves. Laplacian Score~\cite{he2005laplacian}, an unsupervised technique, selects a subset that preserves crucial features of the data. SPEC algorithm~\cite{zhao2007spectral}, an extension of Laplacian Score, is versatile, and suitable for both supervised and unsupervised scenarios, particularly when utilizing the RBF kernel function for measuring data similarity. On the other hand, Fisher Score~\cite{duda2006pattern}, a supervised approach, enhances feature values for similar samples while distinguishing values for dissimilar ones. Trace Ratio Criterion~\cite{nie2010efficient} directly selects the optimal feature subset globally by calculating feature scores.
Information theory-based methods calculate feature importance through metrics such as Information Gain~\cite{lewis1992feature}, which considers features strongly correlated with labels as important for enhanced downstream task performance. Mutual information-based feature selection~\cite{battiti1994using} emphasizes both strong correlation with category labels and low correlation among features. The Minimum Redundancy Maximum Relevance method~\cite{peng2005feature} seeks to combine essential features into the final subset, while FCBF (Fast correlation-based Filter)~\cite{yu2003feature} simultaneously considers feature-class correlation and feature-feature correlation.
Sparse learning-based methods aim to delete unimportant features by minimizing fitting errors and incorporating sparse regularization terms. Feature selection methods based on $\bm{\ell}_{p}$-norm regularization~\cite{tibshirani1996regression, zhu20031} add $\bm{l}_{p}$-norm sparsity penalty terms to downstream task models. $\bm{\ell}_{p,q}$-norm regularization is applied to multi-classification tasks and multiple regression targets, ensuring consistent feature selection across multiple targets. Nonnegative Discriminative Feature Selection (NDFS)~\cite{li2012unsupervised} combines spectral clustering with simultaneous feature selection. Multi-Cluster Feature Selection (MCFS)~\cite{cai2010unsupervised} selects features covering the multi-cluster structure of data without class labels, using spectral analysis to measure the correlation between features.
Statistics-based methods employ statistical measures to evaluate feature correlation without relying on learning algorithms. T-score~\cite{davis1986statistics} utilizes the T-test to assess whether features significantly differentiate between two categories, making it suitable for binary classification problems. Chi-Square Score~\cite{liu1995chi2} evaluates independence between different features and class labels using an independence test. The Gini index~\cite{gini1912variability} quantifies the ability of features to separate instances from different categories. CFS (Correlation-based Feature Subset)~\cite{hall1999feature} employs correlation-based heuristics to evaluate the value of feature subsets.
What's more, there are a lot of deep learning and reinforcement learning based methods.
This work ~\cite{xiao2023beyond} presents a deep learning-based feature selection framework for robust, dimensionality-agnostic optimization.
MARFS ~\cite{liu2023interactive} proposes a Monte Carlo-based feature selection method with efficiency strategies, demonstrating effectiveness through real-world data experiments.
MCRFS ~\cite{liu2021efficient} resents a Monte Carlo based reinforced feature selection method with early stopping and reward-level interactive strategies to improve efficiency and effectiveness in feature selection tasks.
$C^{2}IMUFS$ ~\cite{huang2023imufs} introduces a novel method for incomplete multi-view unsupervised feature selection, using a weighted matrix factorization model and a complementary learning approach for similarity graph reconstruction.
The knockoff filter, creating cost-effective variables replicating the correlation structure of original variables without new data, serves as a negative control for identifying crucial predictors while controlling the false discovery rate, facilitating the integration of potent machine learning tools with robust statistical guarantees.
In this context, Jordon et al.~\cite{jordon2018knockoffgan} tackle the prevalent issue of feature selection in machine learning, introducing a flexible knockoff generation model based on Generative Adversarial Networks.
Sesia et al.~\cite{sesia2019gene} extend the knockoff methodology to scenarios with hidden Markov model-modeled covariate distributions, presenting an exact and efficient algorithm for generating knockoff variables.
Gimenez et al.~\cite{gimenez2019knockoffs} advance the knockoff procedure in statistics to address the challenge of identifying causally relevant features.
Liu et al.~\cite{liu2018auto} address a crucial challenge in the broad application of the Model-X procedure, introducing a model-free knockoff generator that utilizes latent variable representation to approximate feature correlation structure while enabling the identification of important factors and controlling the False Discovery Rate (FDR).

\section{Conclusion}

This article proposes an innovative automated and unsupervised method for the feature selection problem, leveraging the strength of single-agent reinforcement learning, augmented with Knockoff feature generation and matrix reconstruction techniques. By integrating Knockoff information at the pre-training and exploration stages, we've introduced a novel approach to inject pseudo-information, significantly enhancing the learning process. Our framework uniquely constructs an unsupervised reward function using both Knockoff information and matrix reconstruction. 
Through extensive experimentation on a wide array of datasets, our findings validate the robustness and efficacy of the proposed method. Not only does it accelerate the feature selection process, but it also broadens the scope of application for feature selection methods, marking an advancement in the field.
In future work, We aim to enhance the embedding of Knockoff information through more direct methods, such as distributional projections, making the learning process simpler and improving feature selection efficiency and accuracy. Additionally, we will explore the development of more advanced reward mechanisms to enhance the overall effectiveness of our unsupervised feature selection framework.


\bibliographystyle{ACM-Reference-Format}
\bibliography{sample-base}










\end{document}